\documentclass[10pt,twocolumn,letterpaper]{article}

\usepackage{cvpr}
\usepackage{times}
\usepackage{epsfig}
\usepackage{graphicx}
\usepackage{amsmath}
\usepackage{amssymb}

\usepackage{hyperref}
\hypersetup{breaklinks=true,bookmarks=false}

\cvprfinalcopy 

\def\cvprPaperID{****} 

\begin{document}

\title{"You eat with your eyes first": Optimizing Yelp Image Advertising}

\author{Gaurab Banerjee\\
{\tt\small gbanerje@stanford.edu}
\and
Sam Spinner\\
{\tt\small spinners@stanford.edu}\\
\and
Yasmine Mitchell\\
{\tt\small yasminem@stanford.edu}

\and
Stanford University
}

\maketitle

\begin{abstract}
   A business's online, photographic representation can play a crucial role in its success or failure. We use Yelp's image dataset and star-based review system as a measurement of an image's effectiveness in promoting a business. After preprocessing the Yelp dataset, we use transfer learning to train a classifier which accepts Yelp images and predicts star-ratings. Additionally, we then train a GAN to qualitatively investigate the common properties of highly effective images. We achieve 90-98\% accuracy in classifying simplified star ratings for various image categories and observe that images containing blue skies, open surroundings, and many windows are correlated with higher Yelp reviews. 
\end{abstract}

\section{Introduction}
In Q4’19, Yelp had 36 million unique users and seated 30 million diners. It is not a stretch to say that poor Yelp reviews can "kill" a restaurant in today's high-review, high-traffic world. Luca (2016) found that a "one-star increase in Yelp rating leads to a 5-9 percent increase in revenue" with a primary effect for independent restaurants. Furthermore, Luca identified that "consumers do not use all available information and are more responsive to...changes that are...visible" such as images \cite{Luca}. Small businesses tend to operate on razor-thin margins in the food industry and cannot afford the advertising budgets of chain restaurants. On Yelp, a restaurant controls very few aspects of the experience. One such aspect, though, is the images they are able to upload. Business owners currently have no easy way of concretely deciding whether an image is a "good" image to bring in customers. They also do not have a central idea of what a "good" image looks like compared to a "bad" image.

We seek to address the problem identified above by creating a dual set of tools that restaurant owners can use to assess their business. 

First, we build a classifier that assesses the quality of a restaurant's online photographic representation. This classifier will accept restaurant-related images as inputs and predict Yelp star review ratings as an output. A below average image will receive a 1-3.5 star classification, an average image a 4 star classification, and an above average image a 4.5-5 star classification. 

Then, we implement a GAN trained on average and above-average rated images to provide qualitative analysis regarding how business owners can increase the quality of their business's online representation. The images generated by the GAN will capture important features of high-quality images, and can be used as sources of guidance and comparison for business owners.

\section{Related Work}



Recent papers have focused on improving technical accuracy for deep learning related tasks with food. Liu et al (2016) demonstrated the use of a deep convolutional network based on LeNet-5 \cite{lecun1998gradient}, AlexNet \cite{krizhevsky2012imagenet}, and GoogLeNet \cite{szegedy2015going} to show better-than-previous classification accuracy \cite{phoneFood} \cite{liu2016deepfood}. Their approach involved pretraining on ImageNet followed by fine-tuning. Hassannejad et al (2016) used Google's Inception network as inspiration to train on well-known food image datasets: ETH Food-101, UEC FOOD
100, and UEC FOOD 256 \cite{hassannejad2016food}. This group achieved best results for most efficient computation on those datasets at the time. It is clear that classifying on images of food using neural networks is a growing area of research. We believe that the existing literature provides a good backing for this piece of our project.

Generative adversarial networks (GANs) have recently been gaining traction in both academia and the public eye. The now-viral thispersondoesnotexist.com project provides a good demonstration of the capabilities of well-constructed GANs to produce images nearly indistinguishable from real images to the human eye \cite{karras2019analyzing}. Ito (2018) et al demonstrated the use of conditional GANs (cGANs) to produce both recipe and ramen photos \cite{ito2018food}. They showed that "dish discriminator and WGAN-GP are effective for food image domain." One constraint of this study was their main data was restricted to a single, mostly-uniform type of food. Our work can extend this with a dataset comprised of many extremely different dishes without written context labels for training.

\section{Methods}

This project has two parts: 1) classification and 2) GAN.

\subsection{Classification}
\subsubsection{Baseline}
For our baseline, we use ResNet-18 \cite{he2016deep} to do transfer learning and predict star rating. We use the existing model and only change the final fully connected layer to map to our 9 possible star ratings buckets.

\subsubsection{Our Approach}

We use transfer learning as our starting point. Since our dataset is on the order of about 100,000 images sorted to 9 unevenly distributed categories, we tested various hyperparameter combinations, loss functions \cite{janocha2017loss}, optimizers, and whether fine-tuning the ConvNet or using it as a feature extractor is better.

\paragraph{Loss Functions}

It is important to recognize that the 9 classes in our classification are not equidistant. If the true label of a restaurant image is '1.5 stars', the model should be punished less for predicting the label '2 stars' in comparison to predicting the label '5 stars'. Class proximity is a metric we decided to test out. We therefore try distance-based loss such as Mean-Squared Error Loss. MSE loss can be modeled by 
$$
l(x,y) = mean( \{l_1, l_2, ... l_N\}^{\top}), l_n = (x_n-y_n)^2,
$$
where $l(x,y)$ is our loss. In this case, $x$ represents our model outputs while $y$ represents our target values. Due to the nature of this construction, we see that predicted values numerically more distant from the expected value yield higher losses. Our preprocessing to changes the value ranges of our data from 9 classes in [1,5] to 9 classes in [1,10] perform the added value of ensuring all $l_n$ values will be whole numbers since decimal difference won't scale as well with a squaring function.

The second loss function we tested with was Cross-Entropy Loss. This can be modeled by the equation

$$
loss(x,class) = weight[class] \cdot (- log (\frac{exp(x[class])}{\sum_jexp(x[j])}))
$$

where $loss(x,class)$ models the loss per class for each output. For Cross-Entropy Loss, the shape is batch size x \# of classes so the expected target value provided is the index of which class is the correct class for the specific example. Cross-Entropy Loss predicts probability of being in a specific class and calculates log loss based on this. Therefore, there is no preservation of true distance measure from the predicted class to the real class.

\paragraph{Optimizers}

First, we test stochastic gradient descent (SGD) with momentum \cite{keskar2017improving}. Since we used PyTorch\cite{NEURIPS2019_9015}, the SGD with Momentum updates can be modeled by the equations
$$
v_{t+1}= \mu \cdot v_t +g_{t+1}
$$
$$
p_{t+1} = p_t - lr * v_{t+1}
$$
where $p$ denotes the parameters, $g$ is gradient, $v$ is velocity, and $\mu$ is momentum. SGD performs weight updates in the direction that will reduce a mini-batch's error. 

Second, we test Adam \cite{kingma2014adam}. Adam computes to moment estimates and then bias-corrects both. Once these internal metrics are calculated, Adam performs weight updates.
\begin{center}
$m = \beta_1 \cdot m + (1-\beta_1) \cdot (dx)$\\
$m_t = m / (1-(\beta_1)^t)$\\
$v = \beta_2 \cdot v + (1-\beta_2)\cdot((dx)^2)$\\
$v_t = v / (1-(\beta_2)^t)$\\
$x += - learning\_rate \cdot m_t / ((v_t)^{1/2} + eps)$

\end{center}

Adam combines principles of RMSProp and AdaGrad to compute dynamic learning rates for its different parameters. One notable issue with Adam is that it has been known to fail to converge if given the right parameters. 

\paragraph{Hyperparameter Tuning}
We perform hyperparameter tuning on learning rates, learning rate decay, weights, momentum, batch size, etc. by first doing a coarse-grained search on orders of 10 and then doing fine-grained search using random initialization within a defined range.

\paragraph{Fine Tuning vs Feature Extractor}
Fine tuning \cite{peters2019tune} on a pretrained model (in our case ResNet-18), the model is initialized with pretrained weights and then trained normally on our provided images. For feature extraction, the model is initialized with pretrained weights, frozen, and only the final fully connected layer is updated during training.

\paragraph{Model Outputs}
Although we originally trained a model to produce outputs from [1, 10], we found that the difference between adjacent star ratings was somewhat arbitrary and would not be beneficial to a potential user. To improve the usefulness and simplicity of our model, we hypothesize that users see any review [0,3.5] stars as being below average, any review at 4.0 stars as average, and any review [4.5,5] as above average. This is based on our own experience and the relative star distributions in Figure 2. We train a new classifier which accepts the same input images as the baseline but reduces the number of output classes to 3 to reflect these intuitive representations.

\subsection{GAN}

We use a GAN architecture in the generation of our restaurant images. We look at first to the classic GAN framework, which involves a generative model and a discriminator "competing" to decrease their losses.  The generator takes random noise and attempts to create an image that matches the distribution of the true images. The discriminator attempts to predict whether or not an image is a member of the original dataset. Essentially, they are playing a min-max game as follows: 
$$
\min_{G} \max_{D}(D, G) = $$
$$\mathbb{E}_{x\ pdata}[\log (D(x)]
+ \mathbb{E}_{z\ p_z(z)}[\log (1 - D(G(z)))]
$$ 
We will generally follow this classic GAN procedure of training our generator and discriminator in a zero-sum manner. Because we want to produce different classes of images based on an input star rating, we will train different models on the different subsets of real images. \\ 
The principal framework we looked to was StyleGAN2. Both it and its predecessor employ a unique architecture where the input latent code is also transformed into intermediate latent code which allows for the creation of styles and the use of adaptive instance normalization. 


\section{Dataset}

We pull our data from the Yelp Academic Dataset, which is typically used for NLP research but also contains a folder of over 200,000 images. Associating each image with a business and that business's star rating took considerable pre-processing.  Documentation for the original dataset is available \hyperlink{https://www.yelp.com/dataset/documentation/main}{here}. 

\subsection{Attributes}

\begin{figure} [h]
\begin{center}
\fbox{\includegraphics[width = 200px]{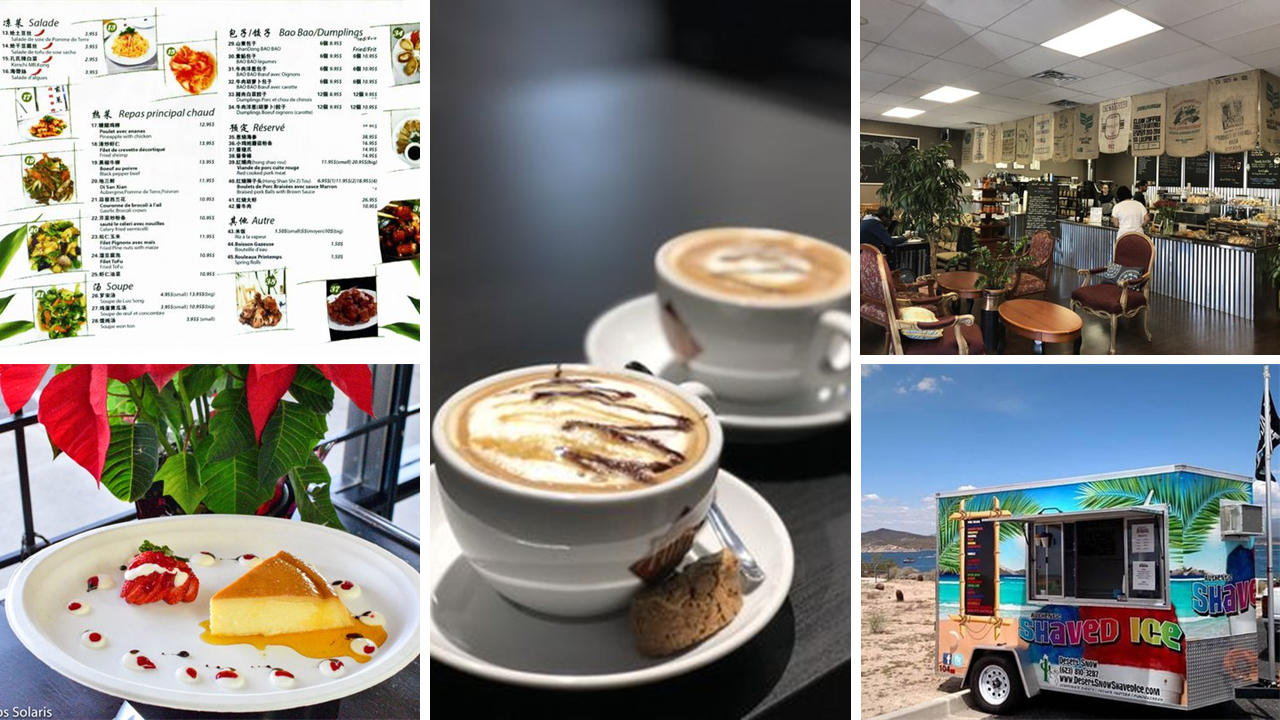}}
\end{center}
  \caption{Sample photos from each of our 5 sub-datasets}
\label{fig:short}
\end{figure}

This particular dataset is comprised of 8,021,122 reviews by 1,968,703 users of 209,393 businesses located in 10 metropolitan areas. Of the many available attributes, one available feature in this dataset is a lightly annotated bank of 200,000 images tied to those businesses.

The metropolitan areas represented are Montreal, Calgary, Toronto, Pittsburgh, Charlotte, Urbana-Champaign, Phoenix, Las Vegas, Madison, and Cleveland.

All of the text-based data is available across several json files (\eg business.json, photo.json, review.json, etc.). Within each json are relevant fields captured and clustered around a couple specific tags: \textit{business\_id} and \textit{user\_id}. The \textit{stars} field contains float star ratings for each business between 0. and 5.0 rounded to the nearest half star.

Importantly, photo.json includes a \textit{label} field for each image which categorizes each image into one of five specific classes: "food", "drink", "menu", "inside" or "outside". Because each category of photos is qualitatively dissimilar from the others, we decided to create 5 different datasets, one for each category.  

Altogether, the dataset is about 20 GB worth of information.

\subsection{Pre-processing}


In order to reduce the feature space size, we select specific fields across multiple json files taken from the Yelp dataset that are of interest to us. We keep \textit{business\_id}, \textit{photo\_id}, \textit{label}, and \textit{stars} only.

For classification, we first map each photo\_ID to a business\_ID, which is in turn mapped to its star rating. Once these two maps are constructed, we use Pillow to open images using their photo\_ID, convert the images to numpy arrays, and finally pad and reduce the images to a constant size. All pre-processed images are thus stored a int8 arrays of dimension (3, 144, 200). The processed image array and star rating are stored together together in a final numpy array that is saved to the disk. This process is repeated for all 5 datasets, resulting in two numpy arrays saved to the disk for each label category. We implement a custom Dataset class which interacts with these saved arrays and is used by a PyTorch DataLoader.

\begin{table}[h]
\begin{center}
 \begin{tabular}{|c c c c|} 
 \hline
  & Train & Val & Test \\ [0.5ex] 
 \hline
 Food & 106,737 & 5,929 & 5,931\\ 
 \hline
 Menu & 755 & 41 & 42 \\
 \hline
 Drink & 11,050 & 613 & 615\\
 \hline
 Inside & 48,415 & 2,689 & 2,691\\
 \hline
 Outside & 12,286 & 682 & 684\\ [1ex] 
 \hline
\end{tabular}
\caption{Number of images in each subset.}
\end{center}
\end{table}

For GAN training, we separated images into new directories by label and star rating, (i.e. one folder that contained all 5-star food images, and another which contained all 2-star inside images). Once these directories were created no further preprocessing was required. 

All of this is encoded in our CustomDataset class.

\subsection{Limitations}

\begin{figure}[h]
\begin{center}
\fbox{\includegraphics[width = 193px]{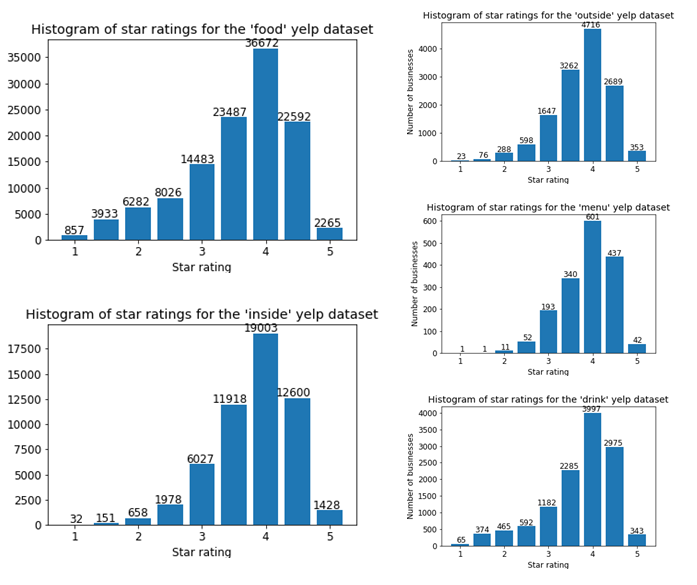}}
\end{center}
  \caption{Histogram of star rating across each label. The star rating for a business was broadcasted to all of its photos.}
\label{fig:short}
\end{figure}

One primary consideration is that we make the explicit choice to assign the same star value to all images from a restaurant. This neglects the likelihood that image quality even within the same business may drastically vary. A second consideration to be made here is that all of the locations represented in our mid-sized cities in North America and therefore any results we obtain will likely not be applicable to preferences in other settings. Another concern is the distribution of ratings. As can be seen in Figure 2, star ratings severely skew left. The distributions are not normal or uniform. Finally, one consideration to keep in mind is that all of these reviews come from Yelp users who only make up a subset of the customer population and their reviews may not necessarily be reflective of any factors outside of their own preferences.


\section{Results \& Discussion}

\begin{figure*}[h]
\begin{center}
\fbox{\includegraphics[width = 450px]{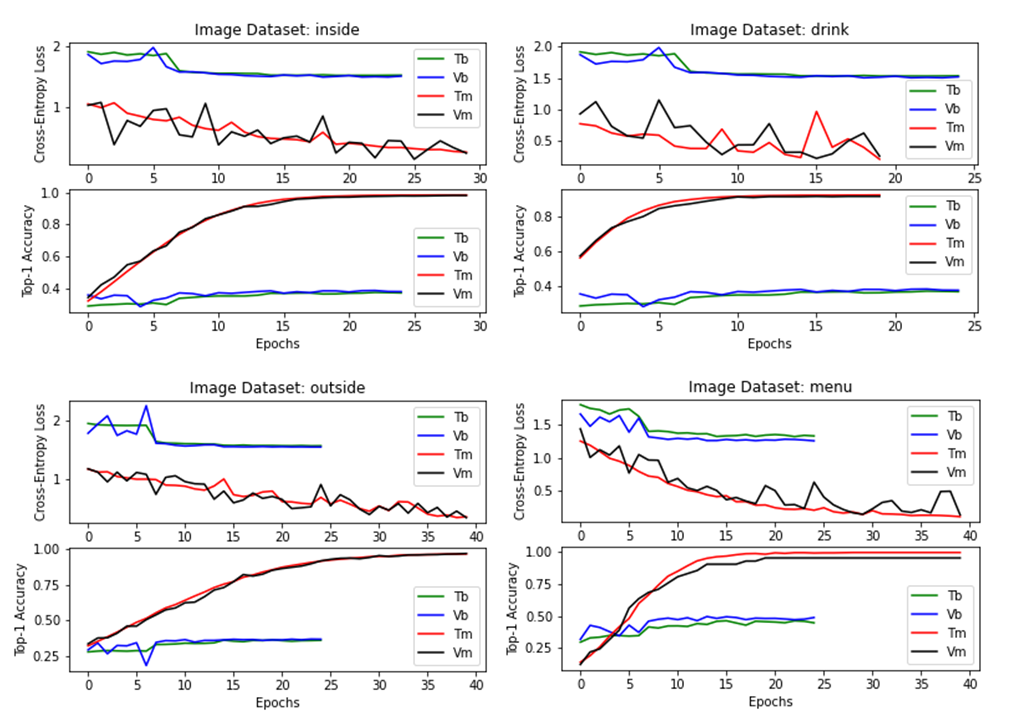}}
\end{center}
  \caption{Our results vs Baseline results for 4 of the datasets. Tb=training baseline, Vb=validaiton baseline, Tm=training with our modifications, Vm=validation with our modifications}
\label{fig:short}
\end{figure*}

\subsection{Classification}

For our baseline of ResNet-18 with a modified final FC layer, Figure 3 displays both Top-1 accuracy and loss. Table 2 displays these metrics as well. Across labels, we see similar trends. The best accuracies demonstrate a considerable improvement from random assignment into 9 classes which would yield an accuracy of roughly 11\%. These values show us that there isn't significant overfitting on our training data. After about 7 epochs, loss and accuracy both converge. Even prior to this point, we see noisy convergence around the final levels so learning rate decay applied at this point appears to be significantly helpful.

\begin{table}[h]
\begin{center}
\title{Baseline Results} \label{tab:title} 
 \begin{tabular}{|c| c c c c c|} 
 \hline
  & Food & \multicolumn{1}{c}{Drink} & Menu   & \multicolumn{1}{c}{Outside} & Inside \\ [0.5ex]
  \hline
 \begin{tabular}[c]{@{}c@{}}Train\\ Loss\end{tabular} &  1.879    & 1.5312 & 1.3647 & 1.5690  & 1.5401 \\
  \hline
\begin{tabular}[c]{@{}c@{}}Val\\ Loss\end{tabular}   &   1.942   & 1.5092 & 1.2613 & 0.3860  & 0.3740 \\
 \hline
\begin{tabular}[c]{@{}c@{}}Train\\ Acc\end{tabular}  &  .3124    & 0.3740 & 0.4338 & 0.3575  & 0.3580 \\
 \hline
\begin{tabular}[c]{@{}c@{}}Val\\ Acc\end{tabular}    &   .3207   & 0.3860 & 0.4952 & 0.3678  & 0.3692\\ [1ex]

 \hline
\end{tabular}

\caption{ResNet-18 results when trained on all 5 datasets across 25 epochs. Acc is Top-1 best accuracy and Loss values are from the epoch with the best val acc value. }
\end{center}
\end{table}

Some of this high accuracy may be able to be attributed to the distrubution of ratings we see in Figure 2. Since the mean rating is clustered at 4 and the ratings are severely skewed left, the model has higher incentive to predict a higher star rating. This effectively could reduce the overall number of classes the model is outputing for. Additionally, we see significantly higher validation accuracy for our \textit{menu} dataset. This could likely be due to it being the smallest dataset and having the least variation. Qualitatively, most menus did not look too different which could resulting in this behavior. Through model checking and verifying outputs, we confirmed that the models were not just outputting a 4 star rating for every image to get to these accuracy levels. The models are outputing different star values for different images.

We notice that there is not significant overfitting or underfitting of the baseline model as the training and validation accuracies track each other fairly well for most of the datasets. Strangely, we see for these datasets that validation tends lower than training loss and validation accuracy tends to be higher than training accuracy. This could arise from the lack of regularization at testing/validation time, as the batch normalization layers in ResNet make use of means and variances that vary from batch to batch.

We tested learning rates from 0.1 to 1e-8 by orders of 10 and then fine-tuned as detailed in our Methods section. In a similar manner, we tested with different momentum values, betas, learning rate decay, learning rate schedules, weights, etc. and found no significant difference beyond convergence time. In our testing with different optimizers, we found no significant difference in best accuracy between Adam and SGD with momentum except that Adam was more noisy at times. MSELoss performed very poorly compared with Cross-Entropy Loss.

\begin{table}[h]
\begin{center}
\title{Best Results} \label{tab:title} 
 \begin{tabular}{|r|r|r|r|r|l|}
\hline
  & Food  & Drink   & Menu   & Outside  & Inside \\ \hline
Train Loss & 0.228 & 0.963 & 1.254  & 1.177  & 1.079  \\ \hline
Val Loss & 0.238 & 1.146 & 1.439 & 1.181 & 1.085  \\ \hline
Test Loss & 0.586 & 0.552 & 0.124 & 0.347 & 0.208  \\ \hline
\multicolumn{1}{|l|}{Train Acc} & \multicolumn{1}{l|}{0.905} & \multicolumn{1}{l|}{0.927} & \multicolumn{1}{l|}{0.993} & \multicolumn{1}{l|}{0.967} & 0.984  \\ \hline
\multicolumn{1}{|l|}{Val Acc}   & \multicolumn{1}{l|}{0.895} & \multicolumn{1}{l|}{0.918} & \multicolumn{1}{l|}{0.951} & \multicolumn{1}{l|}{0.969} & 0.981  \\ \hline
\multicolumn{1}{|l|}{Test Acc}  & \multicolumn{1}{l|}{0.901} & \multicolumn{1}{l|}{0.924} & \multicolumn{1}{l|}{0.976} & \multicolumn{1}{l|}{0.953} & 0.985  \\ \hline
\end{tabular}
\caption{Best results with newly defined buckets.}
\end{center}
\end{table}

We tested to see what would happen when we followed the simplified output approach listed in our Methods section. In this approach, we determined that customers were more sensitive to the perceived "quality group" a restaurant belonged to rather than an individual decimal star rating. We found the bucketed approach brought significantly higher accuracy to our datasets. 

In Figure 3, we can see that the accuracy curves are generally smooth and concave down as we would like to see. Interesting to note is that convergence takes longer. Additionally, loss curves are still generally noisy.

Table 3 holds these results. The relative accuracies of each of our 5 classes is significant. Note that in order of highest to lowest test accuracy we have Inside, Menu, Outsisde, Drink, Food. As we saw in Table 1, our food dataset is approximately twice as large as the next largest dataset and ~150x larger than our Menu dataset. Clearly, size of training data isn't the only metric leading to the differencesin accuracy. We believe this ordering of class accuracy can be attributed to a combination of dataset size and in-class variation. 

The average appearance of a menu is likely far less variable than what a dish in a store may look like. This likely causes menu to have such high accuracy. The insides of most establishment are also likely not too varied but the sheer amount of training examples in this category likely caused this variation. Food and drinks and the outdoor conditions of a restaurant all tend to be more variable. Especially in the food class, very few non-chain restaurants will likely make food that looks the same. In the next section, we report how this may have led to better GAN outputs for certain classes than others.

\subsection{GAN}

\textit{Note: More generated photos can be found in the appendix.}

We created datasets of each class within each star rating (eg all 4.0 star outside images). Then, we trained our GAN using these images.

\begin{figure}[h]
\begin{center}
\fbox{\includegraphics[width = 200px]{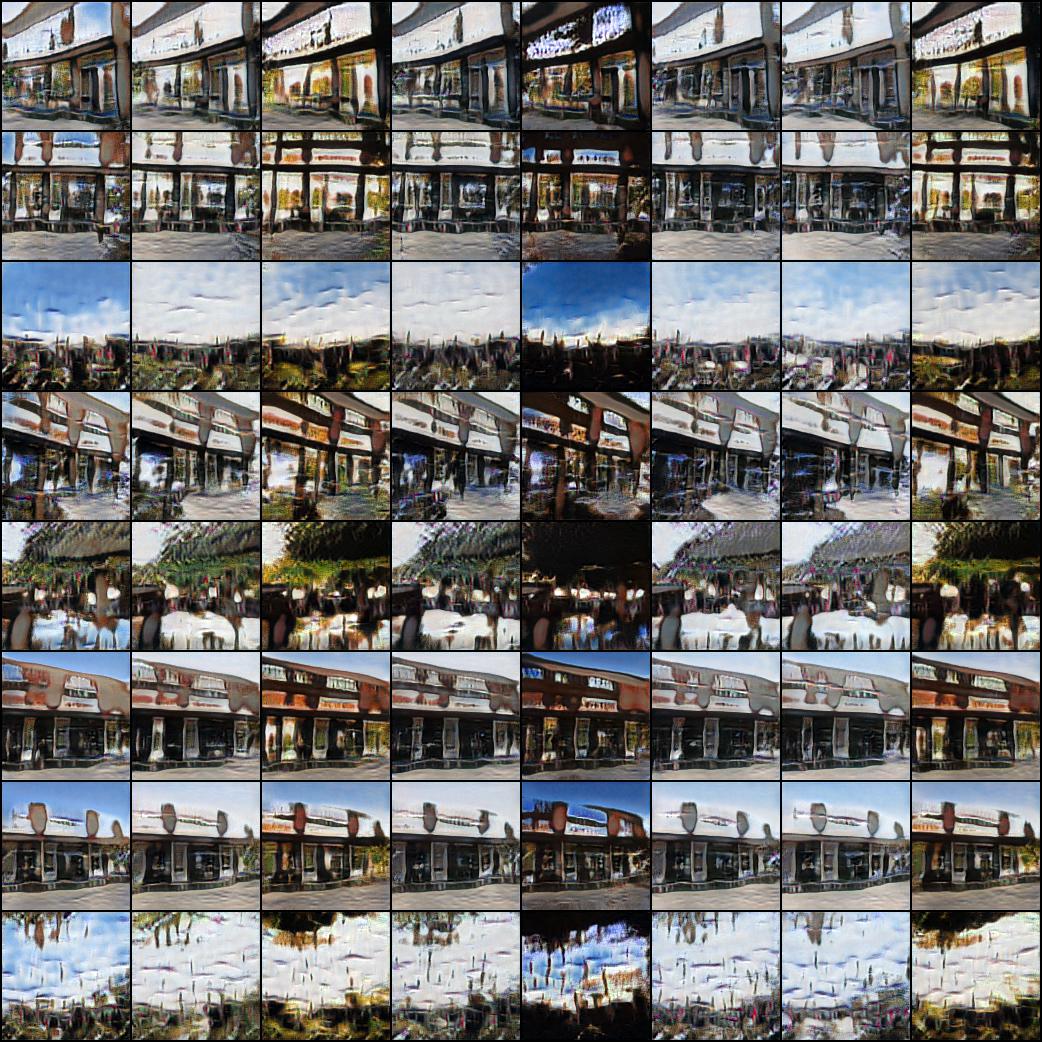}}
\end{center}
  \caption{The image generated by our GAN on its 47th checkpoint of the outside 4.0 images.}
\label{fig:short}
\end{figure}

After training on 4-star outside images, we observe a few specific qualities that are correlated with positive reviews. Large windows, clear blue skies, and clear storefronts are prevalent. Additionally, the generated images display storefronts in conjunction with geographical features or landmarks in view. This suggests that restaurant location and the surrounding environment ambience are important to consumers. We also notice that restaurants in other generated photos are placed in a natural outdoor setting. Although business owners might be limited on their location, beautifying the restaurant's surroundings with natural elements would likely increase attractiveness. 

\begin{figure}[h]
\begin{center}
\fbox{\includegraphics[width = 100px]{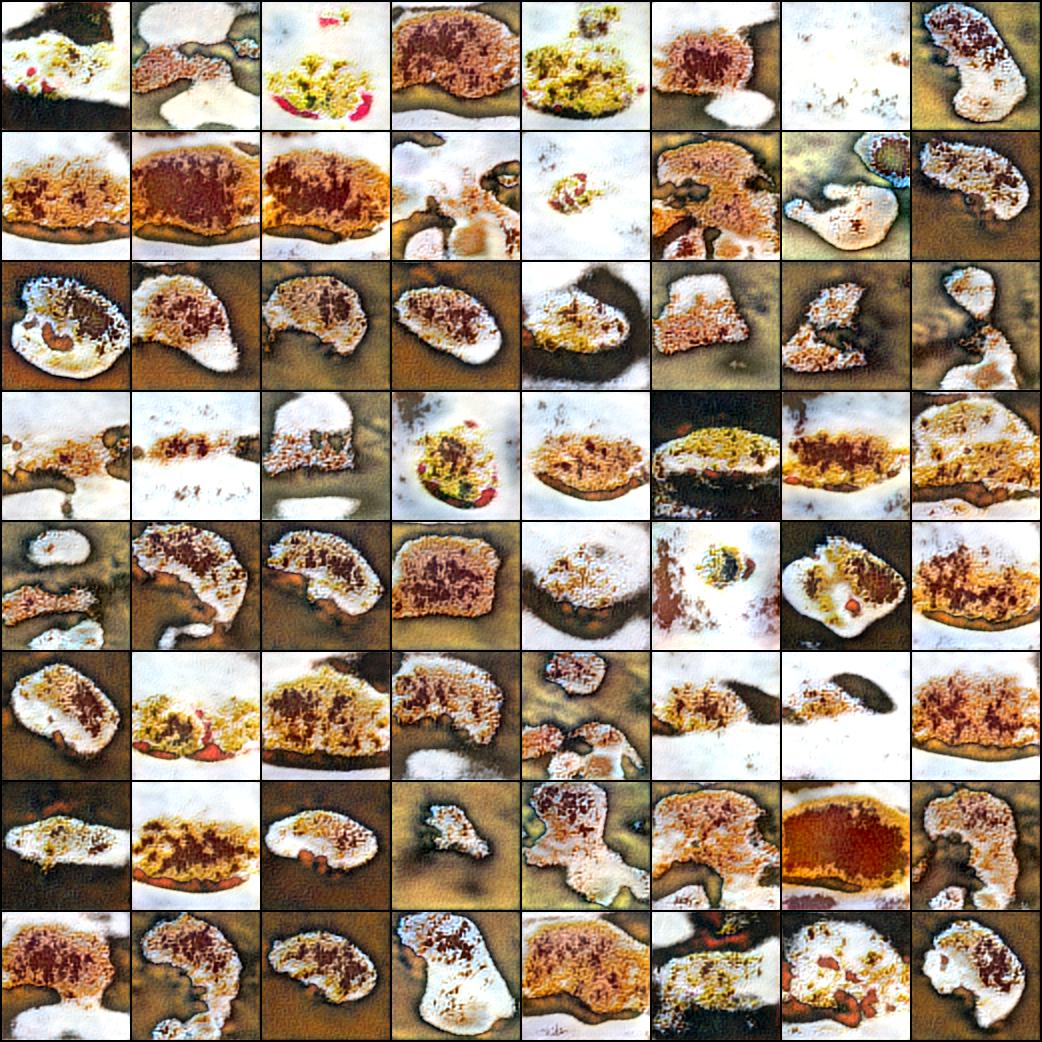}}
\fbox{\includegraphics[width = 100px]{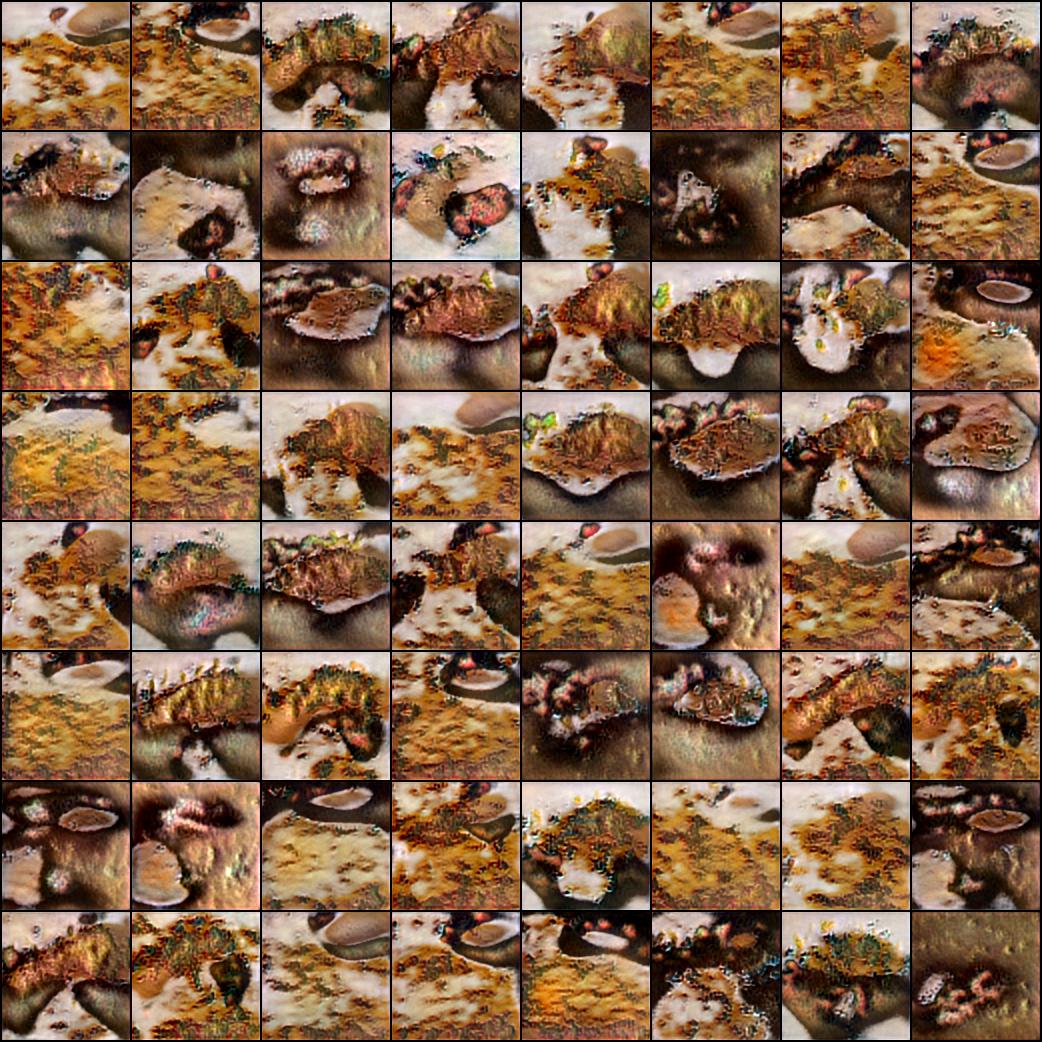}}
\fbox{\includegraphics[width = 100px]{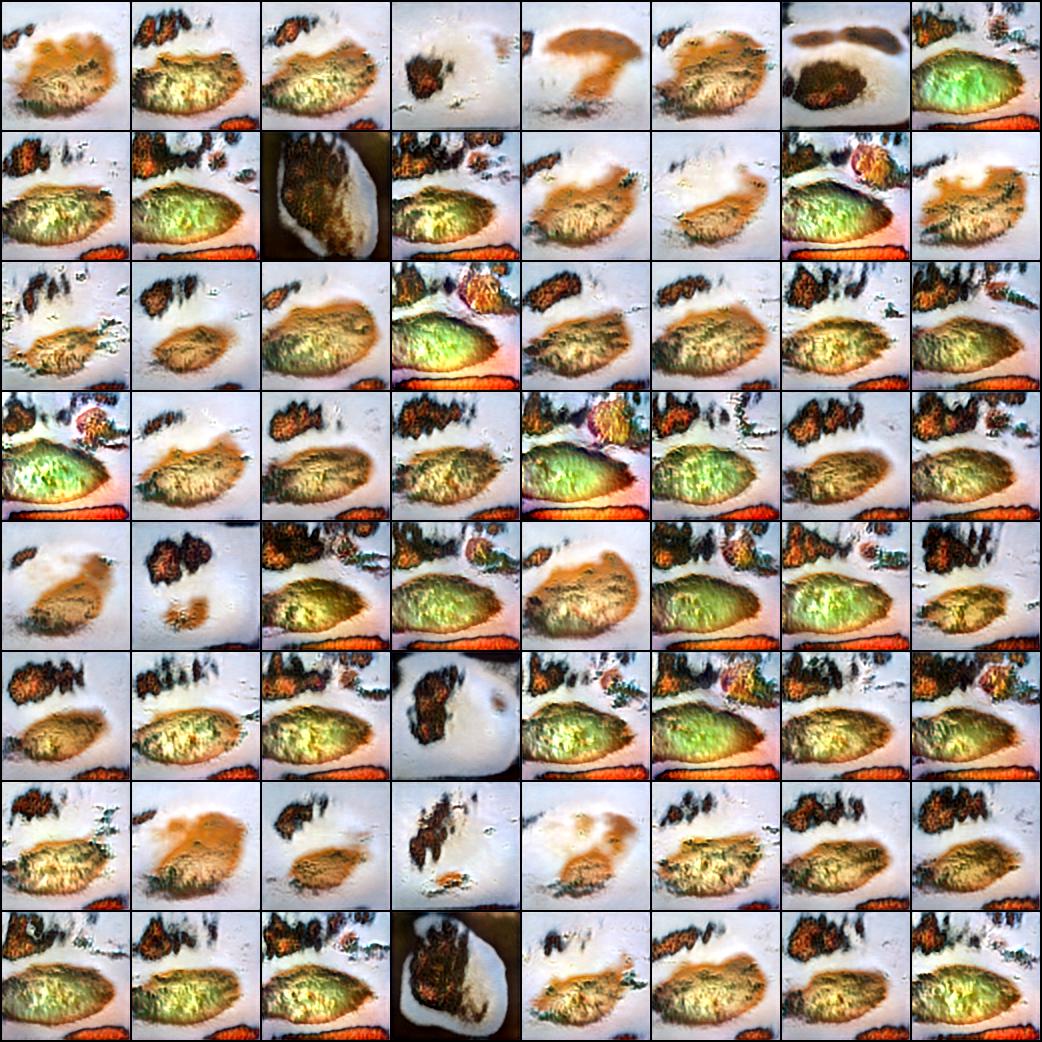}}
\fbox{\includegraphics[width = 100px]{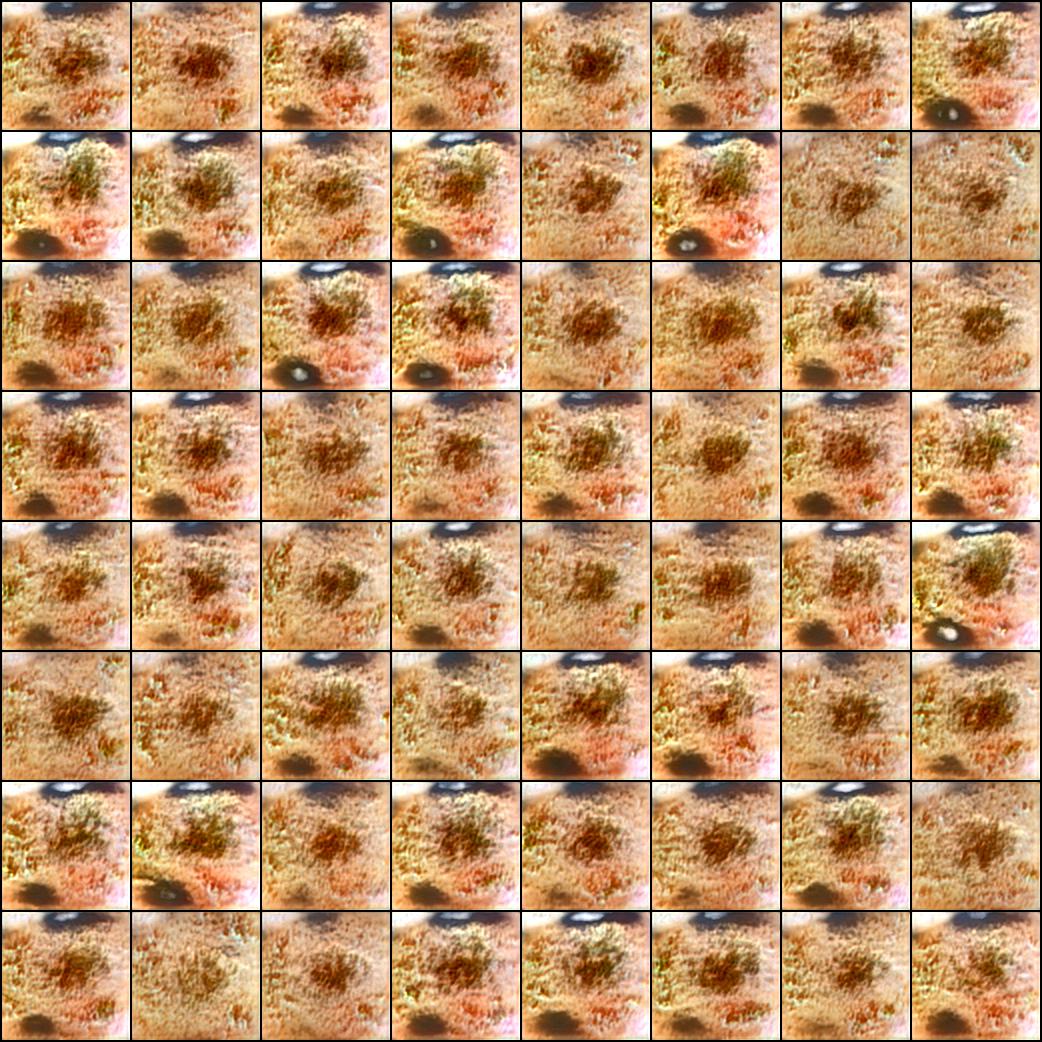}}
\end{center}
  \caption{Checkpoint number generated at from left to right, then top down: 7, 15, 27, 45 (5.0 Food Images)}
\label{fig:short}
\end{figure}

In Figure 5, we can see that food images did not perform well when generated. Again, we attribute this to the great variation in food appearance and style not just between cuisine and restaurant, but fundamentally to how dishes of food look. We believe far better quality outputs may be generated if the dataset is more significantly divided along some relevant features. \\
In general, we saw a trend across all sets of images where the both the visual quality and generator accuracy of model-generated images would plateau and then decrease. We see an example of this in checkpoint 56 in Figure 8. We predict this was caused by the small size of the datasets, as each type-star rating subset only had a few thousand images a piece.

\section{Conclusion}


In this project, we created a set of tools to assist business owners in boosting their Yelp reviews through their image advertising. Our classifiers achieved >90\% accuracy on menu, drink, inside, and outside pictures using our intuitive reduced-class approach. For our best results, we used Cross-Entropy Loss, SGD with momentum and batch sizes of 16. The optimal learning rates for each category were: 1e-7, 1e-6, 2.5e-8, 1e-8, and 1e-8 for drink, menu, outside, inside, and food, respectively.   

Our GAN outputs enabled us to visualize key potential indicators of a great photo in each of these categories. These qualitative results can be utilized to stage advertising photos to appeal to consumer tastes.

\begin{figure}[h]
\begin{center}
\fbox{\includegraphics[width = 200px]{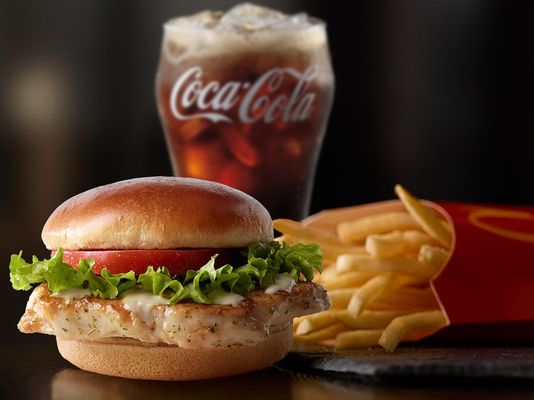}}
\end{center}
  \caption{Example of a 1-star rated restaurant's food image}
\label{fig:short}
\end{figure}

At this point, we wish to explore higher fidelity generated images. Additionally, we think a further breakdown by cuisine or meal type is necessary. Figure 6 shows an objectively well-styled (AKA professionally created) food image. However, this image was rated by customers as from a 1-star restaurant. The image is clearly an advertising photo and not exactly what was served in the restaurant. We believe there can be many confounding variables here and so many questions are raised? Are fast-food or "cheaper" restaurants inherently lower-rated? Do customers prefer real photos to advertising photos? Are certain cuisines higher rated than others?

We also wish to explore the customer perception side of advertising. Why are ratings so severely skewed left? Does photo captioning contribute to star rating and, if so, how? Do customers actually care about the images or is this correlation we identified not actually involved in restaurant perception?

\section{Appendix}

The contents of this section can be found on pages 8-9.

\begin{figure*}[h]
\begin{center}
\fbox{\includegraphics[width = 200px]{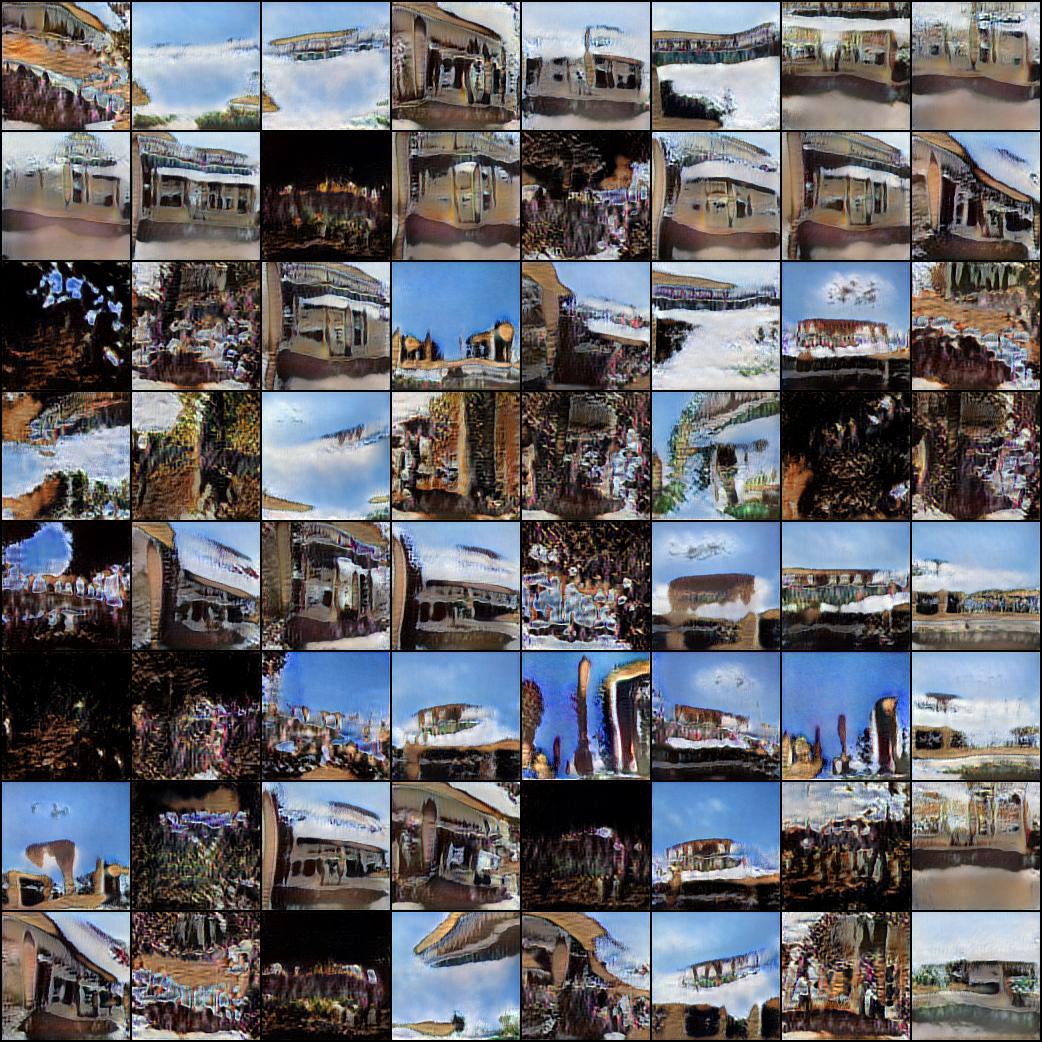}}
\fbox{\includegraphics[width = 200px]{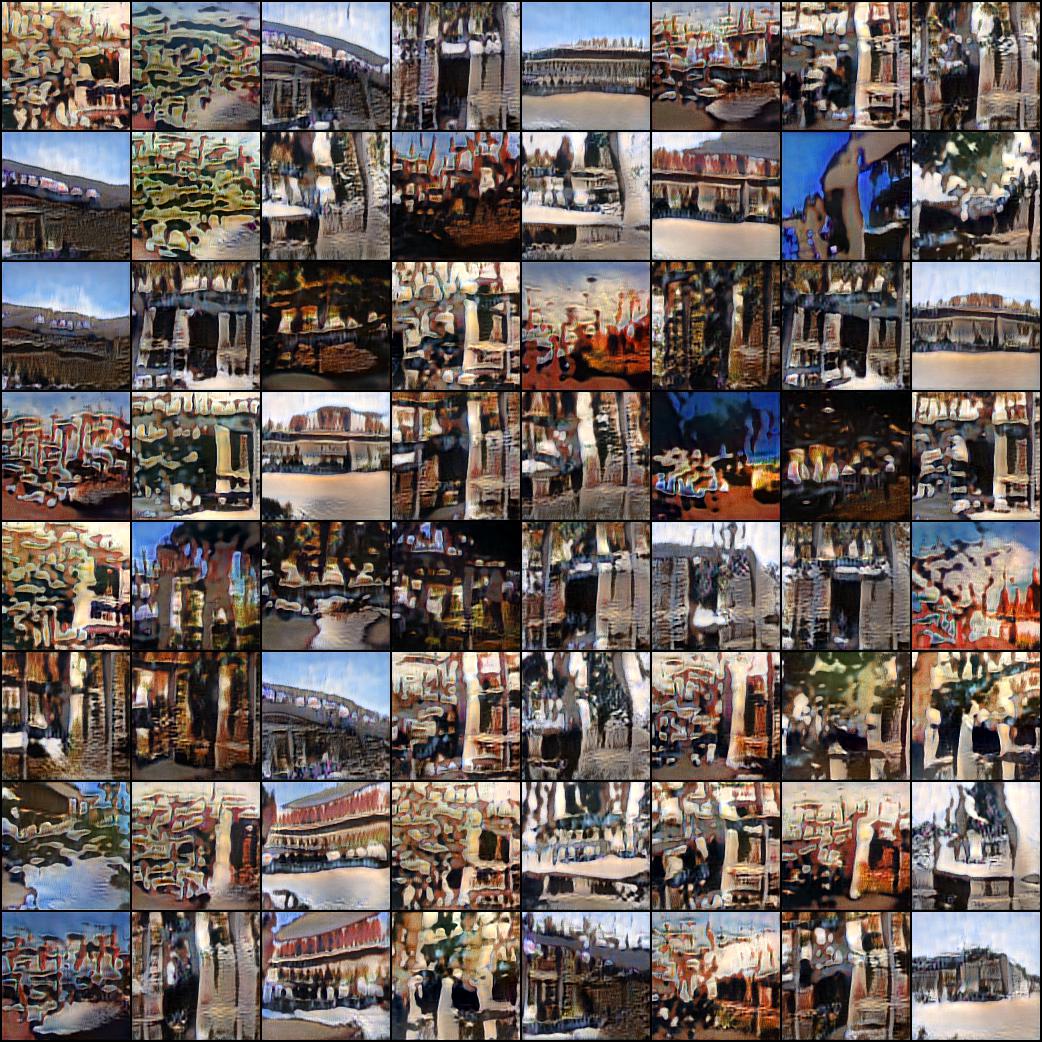}}
\fbox{\includegraphics[width = 200px]{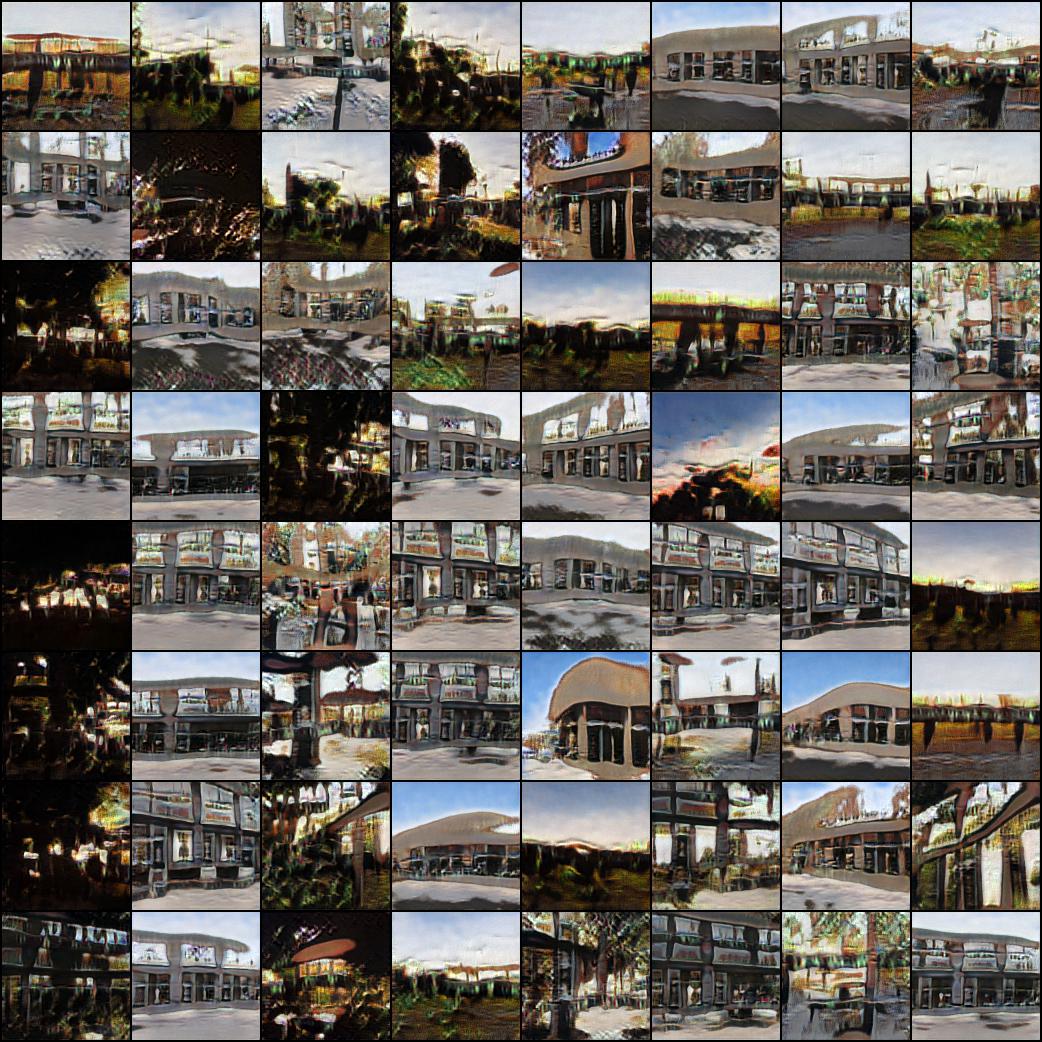}}
\fbox{\includegraphics[width = 200px]{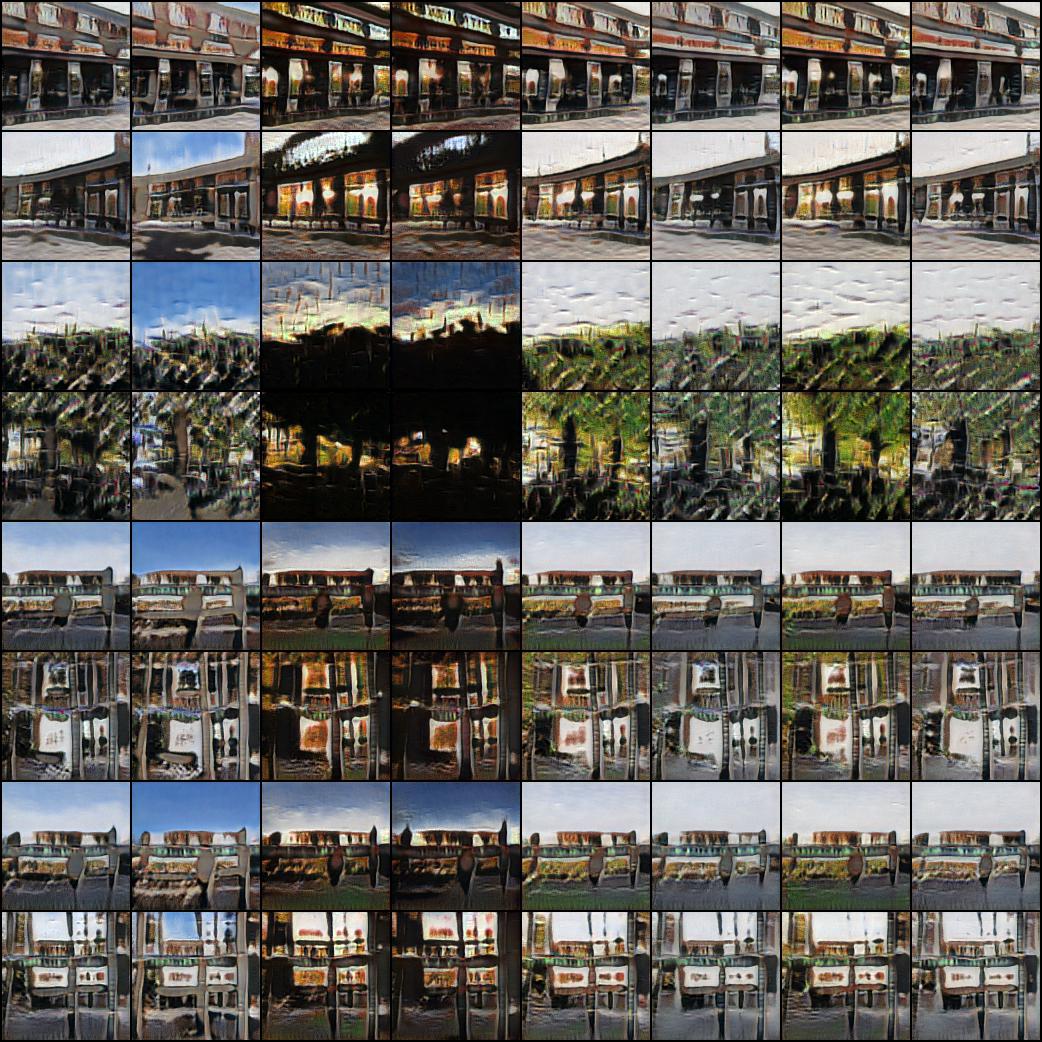}}
\end{center}
  \caption{Images generated at checkpoints 23, 27, 41, and 49 for 4.0 Outside Images from left to right, top down}
\label{fig:short}
\end{figure*}

\begin{figure*}[h]
\begin{center}
\fbox{\includegraphics[width = 100px]{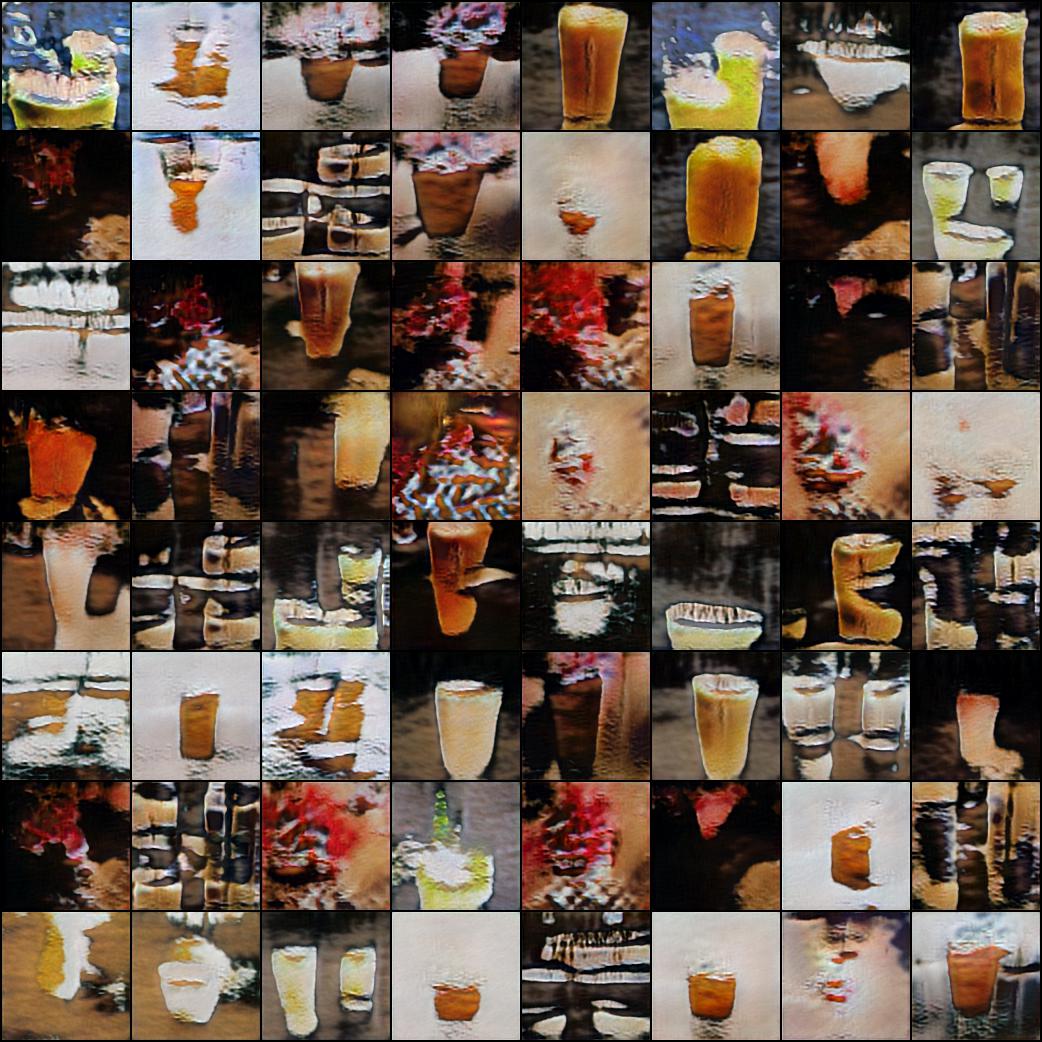}}
\fbox{\includegraphics[width = 100px]{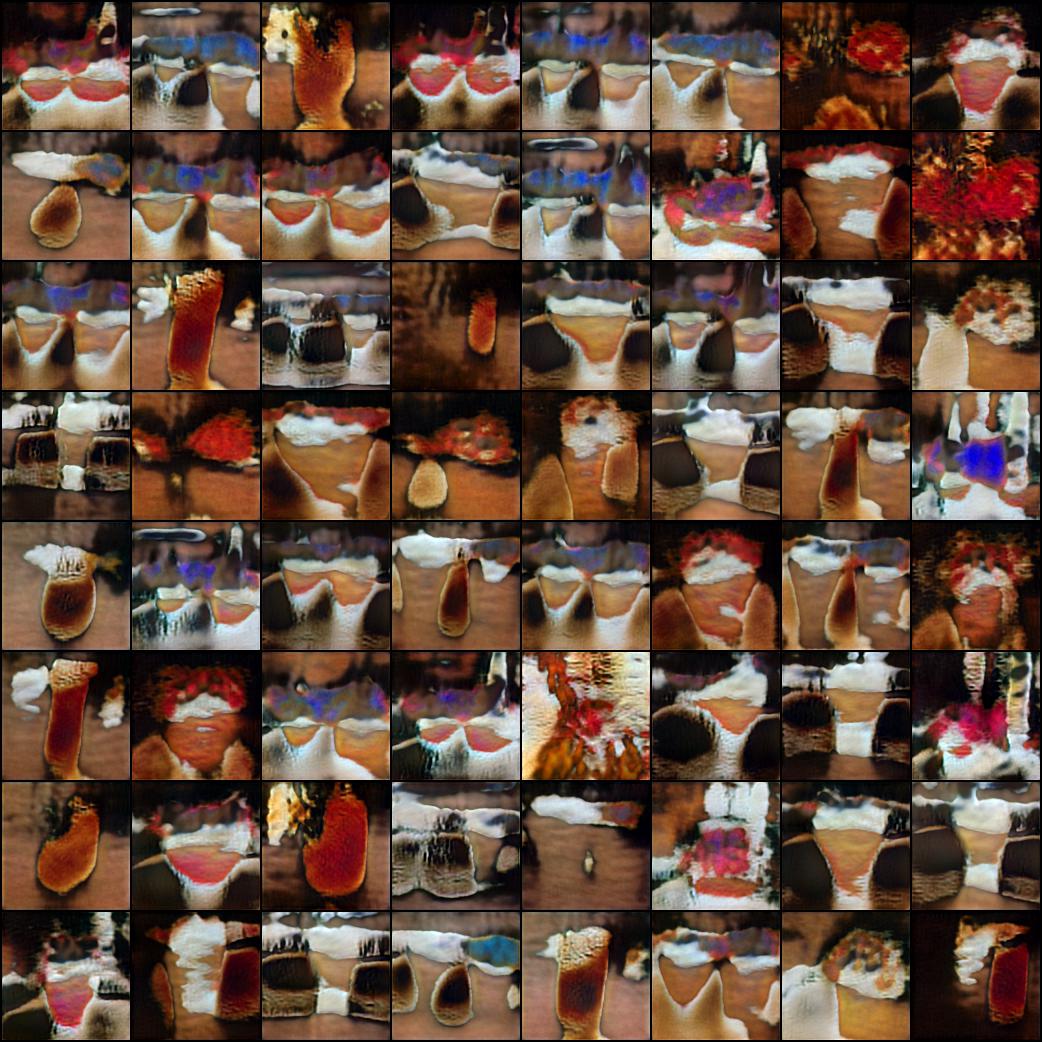}}
\fbox{\includegraphics[width = 100px]{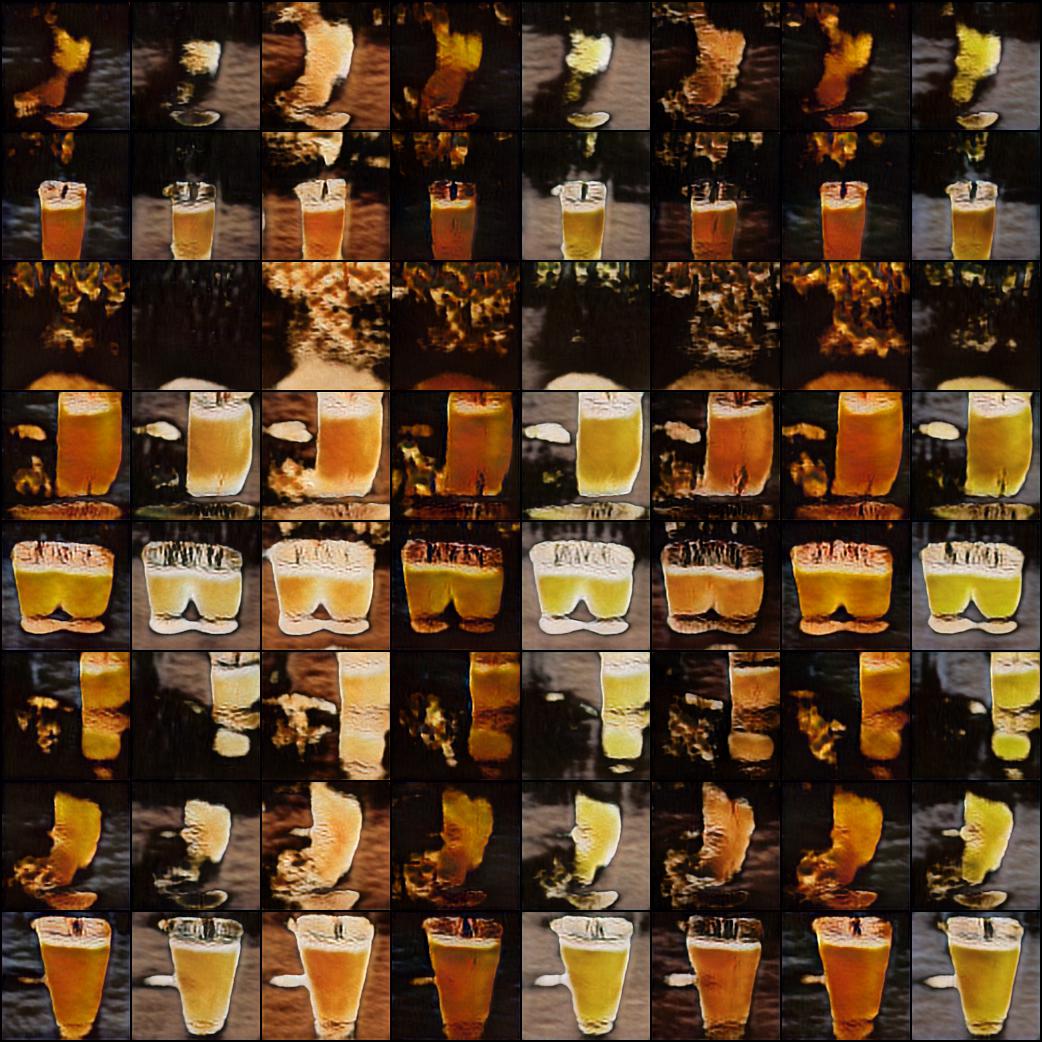}}
\fbox{\includegraphics[width = 100px]{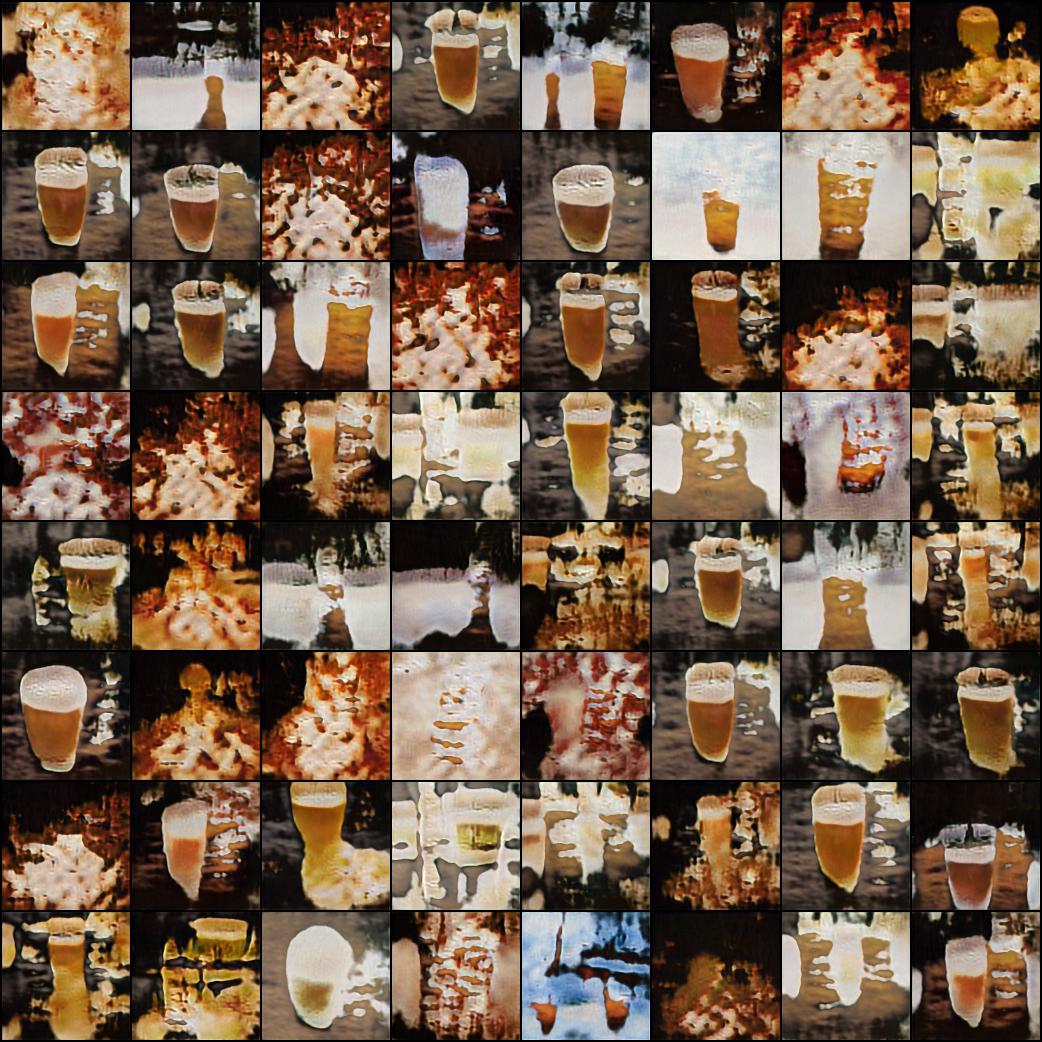}}
\fbox{\includegraphics[width = 200px]{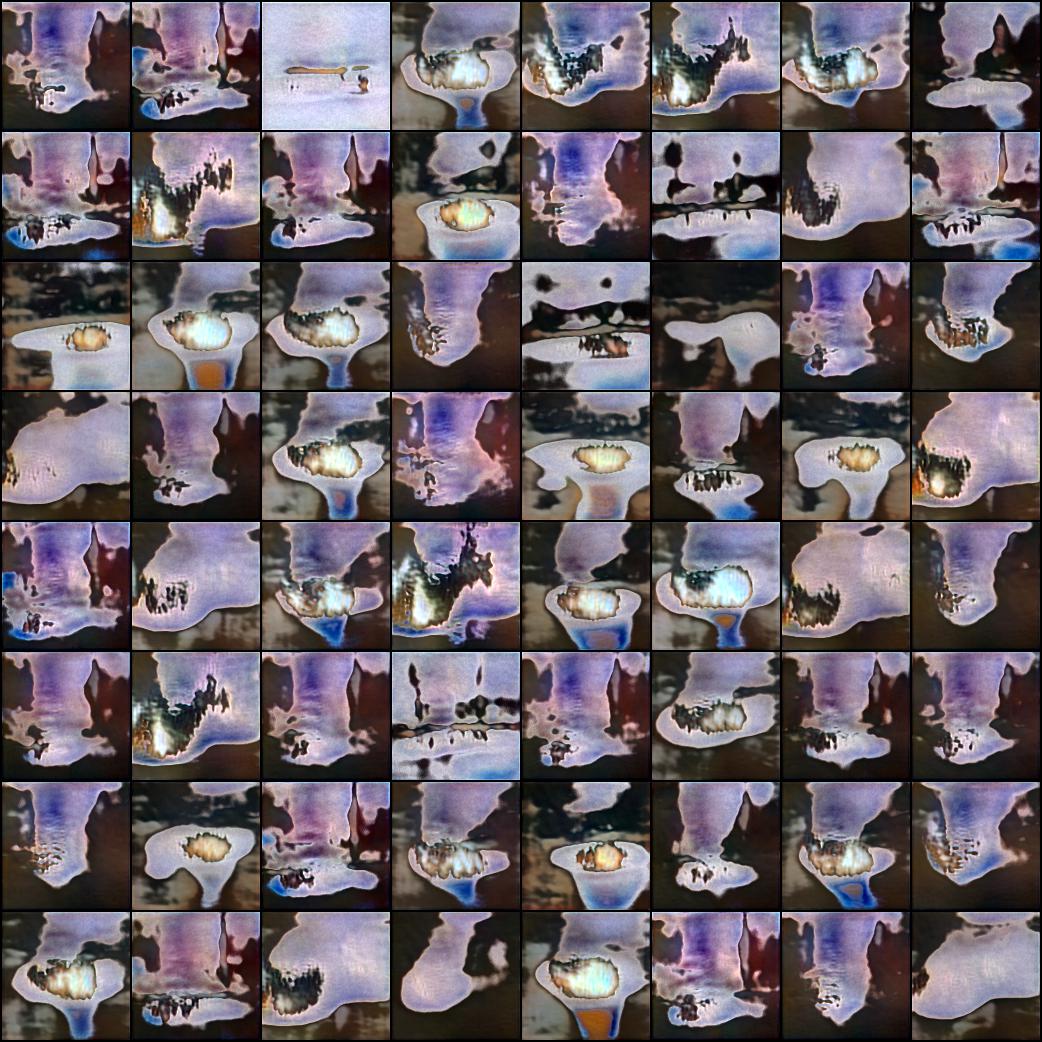}}
\fbox{\includegraphics[width = 200px]{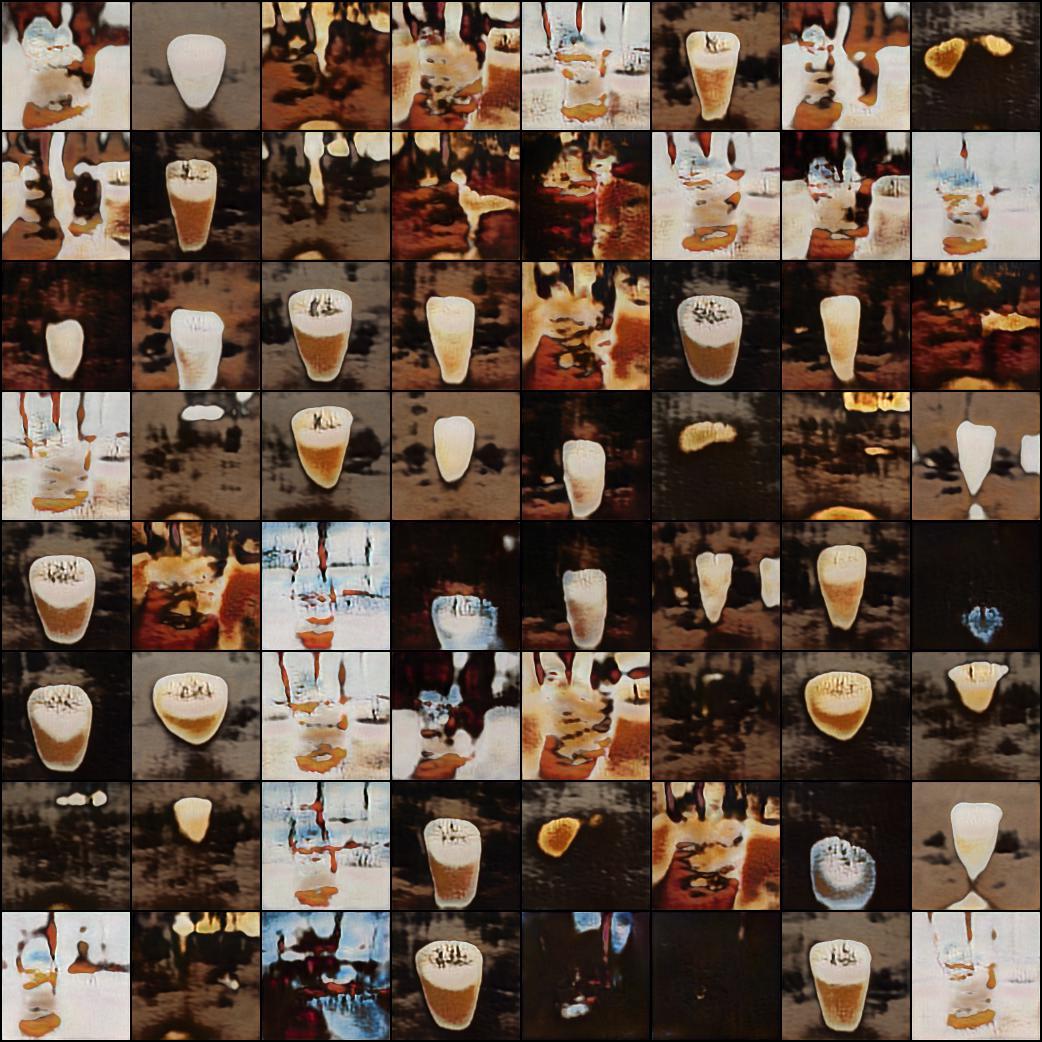}}
\end{center}
  \caption{Checkpoint number generated at from left to right, then top down: 23, 24, 30, 41, 56, 57 (4.0 Drink Images)}
\label{fig:short}
\end{figure*}

\section{Contributions \& Acknowledgements}
GB worked on classification, baseline, and visualization work. SS worked on preprocessing, baseline, and classification. YM worked on GANs. All authors contributed to the writeup.

{\small
\bibliographystyle{ieee}
\bibliography{egbib}
}

\end{document}